\documentclass{article}
\usepackage{ijcai19}

\usepackage{times}
\usepackage{soul}
\usepackage{url}
\usepackage{hyperref}
\usepackage[utf8]{inputenc}
\usepackage[small]{caption}
\usepackage{graphicx}
\usepackage{amsmath}
\usepackage{booktabs}
\usepackage{algorithm}
\usepackage{algorithmic}
\usepackage{times}
\usepackage{epsfig}
\usepackage{graphicx}
\usepackage{amsmath}

\usepackage{amssymb}
\usepackage{etoolbox}

\newcommand{\MyMapTemplatePrefixc}[4]{\expandafter#1\csname#3#4\endcsname{#2{#4}}} 
\forcsvlist{\MyMapTemplatePrefixc {\def} {\mathcal}{c}} {A,B,C,D,E,F,G,H,I,J,K,L,M,N,O,P,Q,R,S,T,U,V,W,X,Y,Z}  

\newcommand{\MyMapTemplatePrefixtb}[5]{\expandafter#1\csname#4#5\endcsname{#2{#3{#5}}}} 
\forcsvlist{\MyMapTemplatePrefixtb {\def} {\tilde}{\mathbf}{t}} {A,B,C,D,E,F,G,H,I,J,K,L,M,N,O,P,Q,R,S,T,U,V,W,X,Y,Z}  
\forcsvlist{\MyMapTemplatePrefixtb {\def} {\tilde}{\mathbf}{t}} {0,1,a,b,c,d,e,f,g,h,i,j,k,l,m,n,o,p,q,r,s,u,v,w,x,y,z}  

\newcommand{\MyMapTemplateNoPrefix}[3]{\expandafter#1\csname#3\endcsname{#2{#3}}}
\forcsvlist{\MyMapTemplateNoPrefix {\def} {\mathbf} } {0,1,a,b,c,d,e, f, g, h, i, j, k, l, m, n, o, p, q, r, u, v, w, x, y, z} 
\forcsvlist{\MyMapTemplateNoPrefix {\def} {\mathbf} } {A,B,C,D,E,F,G,H,I,J,K,L,M,N,O,P,Q,R,S,T,U,V,W,X,Y,Z}  

\newcommand{\ie}{{i.e.}}
\newcommand{\eg}{{e.g.}}
\newcommand{\resp}{{resp.}}

\usepackage{xcolor}

\title{Known-class Aware Self-ensemble for Open Set Domain Adaptation}

\author{
	Qing Lian$^1$\and
	Wen Li$^2$\and
	Lin Chen$^3$\And
	Lixin Duan$^1$\\
	\affiliations
	$^1$Big Data Research Center, University of Electronic Science and Technology of
	China\\
	$^2$EHT Zurich
	$^3$Huawei Technologies\\
	\emails
	\{lianqing1997, gggchenlin, lxduan\}@gmail.com,
	liwen@vision.ee.ethz.ch,
}

\begin{document}
\maketitle

\begin{abstract}
Existing domain adaptation methods generally assume different domains have the identical label space, which is quite restrict for real world applications. In this paper, we focus on a more realistic and challenging case of open set domain adaptation. Particularly, in open set domain adaptation, we allow the classes from the source and target domains to be partially overlapped. In this case, the assumption of conventional distribution alignment does not hold anymore, due to the different label spaces in two domains. To tackle this challenge, we propose a new approach coined as Known-class Aware Self-Ensemble (KASE), which is built upon the recently developed self-ensemble model. In KASE, we first introduce a Known-class Aware Recognition (KAR) module to identify the known and unknown classes from the target domain, which is achieved by encouraging a low cross-entropy for known classes and a high entropy based on the source data from the unknown class. Then, we develop a Known-class Aware Adaptation (KAA) module to better adapt from the source domain to the target by reweighing the adaptation loss based on the likeliness to belong to known classes of unlabeled target samples as predicted by KAR. 
Extensive experiments on multiple benchmark datasets demonstrate the effectiveness of our approach.
\end{abstract}

\section{Introduction}
Open set domain adaptation has been drawing increasing attention from the computer vision community in recent years~\cite{Busto_2017_ICCV,Saito_2018_ECCV,Peng2018Syn2RealAN}. Different from the conventional domain adaptation, where the source and target domains are assumed to contain exactly the same object classes (a.k.a., closed set domain adaptation), open set domain adaptation tackles a more realistic scenario, where only a few classes of interest are shared by the source and target domains, and the remaining ones in source and target domains are totally different.

\begin{figure*}[t]
\centering
\includegraphics[width=0.8\linewidth]{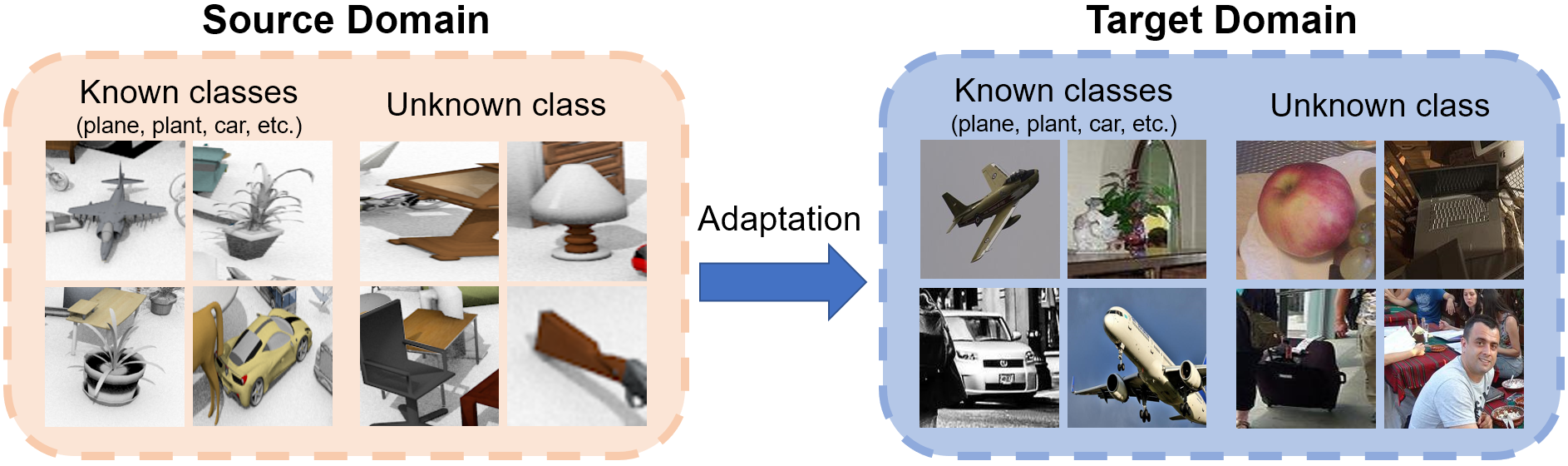}
\caption{Illustration of open set domain adaptation. It is assumed that source and target domains both contain images from the same set of classes (i.e., known classes). There are also unknown classes that exist in both domains, but those classes in the source domain do not overlap with the ones in the target domain.}\label{fig:challenge}
\vspace{-5mm}
\end{figure*}

For closed set domain adaptation, many existing methods proposed to reduce the distribution mismatch between the source and target domains by using different measurements (e.g., such as Maximum Mmean Discrepancy (MMD) and $\mathcal{A}$-distance)~\cite{dan,pmlr-v37-ganin15,jan}. However, such an assumption may not hold any more in the open set domain adaptation scenario, as those measurements could be badly affected by the data coming from those different and unknown classes in the two domains and lead to poor estimation of domain mismatch. As a result, negative transfer may occur and bring down the adaptation performance. To that end, people have developed a few approaches to handle the open set task. \cite{Busto_2017_ICCV} proposed the pioneering work which aims to learn to map from the source samples to a subset of the target, based on the known category information. And \cite{Saito_2018_ECCV} relied on adversarial learning to separate known and unknown samples based on a deep learning framework. Very recently, \cite{baktashmotlagh2018learning} tried to find a subspace based on samples from known classes through learning a factorized representations.

Since the information of known (a.k.a., shared) classes from the source and target domains play a very important role for the open set scenario as shown in the previous work~\cite{Saito_2018_ECCV,baktashmotlagh2018learning}, in this work we focus on how to effectively utilize labeled source data from the known classes to i) identify how likely a target sample comes from a known or unknown category, and to ii) adapt the source knowledge to the target domain based on the unlabeled target training samples.

We realize the above two aspects based on the recently proposed self-ensemble~\cite{french2018selfensembling} approach, owing to its good performance for closed set domain adaptation. We thus coin our method KASE, for Known-class Aware Self-Ensemble. KASE improves Self-Ensemble with two new modules, Known-class Aware Recognition (KAR) and Known-class Aware Adaptation (KAA), to deal with two aspects, respectively. On one hand, in the KAR module, we propose to minimize the cross-entropy loss for the known classes while maximizing the entropy loss for the unknown class in the source domain. This is motivated from the intuition that a good domain adaptation model would achieve better performance (in other words, a lower classification loss) on the known classes in both domains, but might get worse (or a higher classification loss) on an unknown class. On the other hand, in the KAA module, we only use the samples from the shared known classes to perform domain adaptation, such that the adapted model should well classify the target samples from those known classes. Specifically, we improve the sample reweighting strategy in the SE from binary weight to the continuous weights based on how likely an unlabeled target sample belongs to known classes predicted from the KAR module. Based on such reweighting strategy in the teacher-student model difference part, the target known-class sample would play a more important role and the negative effect brought from the target unknown-class could be alleviate in adapting the model.  
We evaluate our KASE through extensive experiments on three benchmark datasets: Syn2Real-O~\cite{Peng2018Syn2RealAN} Office-31~\cite{Saenko_office} and Digits. The superior performance of KASE demonstrates the effectiveness of the KAR and KAA modules for open set domain adaptation.

The remainder of this paper is organized as follows. Section~\ref{sec:related-work} reviews related literature. Section~\ref{sec:kase} introduces the detailed methodology of our proposed KASE method. Moreover, Section~\ref{sec:experiment} presents the experiments, and Section~\ref{sec:conclusion} draws conclusive remarks.

\section{Related Work}\label{sec:related-work}
\noindent{\bf Open set recognition.} Open set recognition is the topic of addressing the class label space mismatch between the training and test data. Recently, several works are contributed to this topic. ~\cite{os_svm} proposed to assign a probability confidence threshold based on the open space risk to reject the unknown classes. ~\cite{openmax} utilized the activation vector to estimate the network failure and proposed a OpenMax layer to detect the unknown classes. ~\cite{g_openmax} further improveed the OpenMax by utilizing the GAN to explicit the probability over unknown categories.

\vspace{3pt}
\noindent{\bf Domain adaptation.} Domain adaptation aims at reducing the annotation burdens for a particular learning task through transferring off-the-shelf knowledge from related source domains~\cite{pan2010survey}. Basically, the main challenge in domain adaptation is the domain shift between the source domain and the target domain, which heavily affects the classifier's cross-domain generalization ability~\cite{ben2010theory}. Among the recent works, maximum mean discrepancy (MMD)~\cite{dan,tzeng2014deep,jan} and domain adversarial training~\cite{pmlr-v37-ganin15,tzeng2017adversarial} are the most two common techniques for domain match. Additionally, generative adversarial networks (GANs) are widely leveraged to alleviate the domain mismatch in the pixel level~\cite{taigman2016unsupervised,bousmalis2017unsupervised}, which is equivalent . Over the past, proxy-label mechanism is also becoming increasingly practical for domain adaptation. In~\cite{saito2017asymmetric}, tri-training strategy is explored to assign pseudo labels to target images, which are then used to train the final classifiers. In~\cite{french2018selfensembling}, French \emph{at al.} leveraged a self-ensemble teacher network to produce pseudo assignments on the target images and pushed the student network to act like the teacher net.

\vspace{3pt}
\noindent{\bf Open set domain adaptation.} However, conventional domain adpataion works generally assumed that the source and the target domains share identical categories, which may not hold in real world application. Currently, there are two kinds of ways to address this problem with different views: 1) Partial domain adaptation; 2) Open set domain adaptation. In this work, we focus on open set domain adaptation, where the target domain could contain classes that not present in the source domain~\cite{Busto_2017_ICCV}. Several works have been proposed recently to address this issue. For example, \cite{Busto_2017_ICCV} explicitly treated the outliers as a particular category and perform knowledge transfer in a self-taught fashion, while \cite{Saito_2018_ECCV} proposed to detect target outliers through a variant of domain adversarial training that allows the feature generator to reject target images as outliers instead of aligning them with the source samples from the known categories. Furthermore, \cite{baktashmotlagh2018learning} proposed a framework to factorizes the data into shared and private sub-spaces and encourage discriminability over the shared representations. In partial domain adaptation, ~\cite{Cao_2018_CVPR,Cao_2018_ECCV} proposed to alleviate the negative effect from the extra labeled source data by reweighting the source sample based on the classification probability for all classes. However, such reweighting strategy could not be applied in open set domain adaptation where the target domain label space is not the subset of the source domain.

In this work, we propose to address the open set domain adaptation with a new solution. We build our approach upon the recent proposed self-ensemble model, which has shown promising performance on closed set domain adaptation tasks. Unlike traditional distribution alignment methods, the teacher-student networks design makes it naturally less sensitive to the category changes across domains. We further improve the self-ensemble model for open set domain adaptation with our newly developed known-class aware recognition and known-class aware adaptation modules to effectively address the impact of the unknown class in the target domain. Extensive experiments validates the superiority of our proposed model over existing closed set and open set domain adaptation approaches.

\begin{figure*}[t]
\centering
\includegraphics[width=0.9\textwidth]{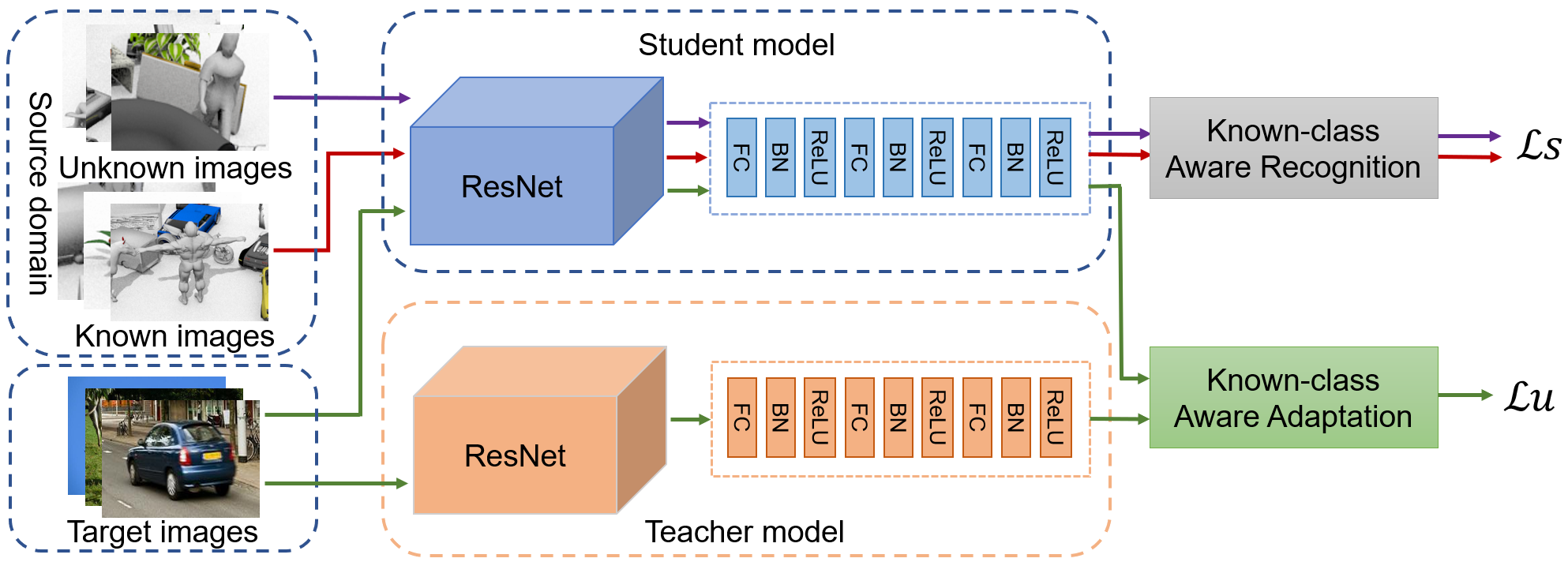}
\vspace{-5pt}
\caption{Overview of our proposed KASE method.}\label{fig:model}
\vspace{-5mm}
\end{figure*}

\section{Known-class Aware Self-Ensemble (KASE)}\label{sec:kase}
In this section, we present details of our proposed KASE method for open set domain adaptation.

Formally, in open set domain adaptation, the training data consists of a labeled source domain and an unlabeled target domain. We denote the source domain as $\{(\x^s_i, y^s_i)|_{i=1}^{n_s}\}$ where $\x^s_i$ and $y^s_i$ are the $i$-th sample and its label, respectively. 
Similarly, the target domain can be denoted as $\{\x^t_i|_{i=1}^{n_t}\}$, where $\x^t_i$ is an unlabeled sample.  


Denoting $\cY_s$ (\resp, $\cY_t$) as the label space of source (\resp, target) domain samples, the traditional closed set domain adaptation assumes $\cY_s = \cY_t$, while in open set domain adaptation it does not hold. We refer to their common categories $\cY_s \cap \cY_t$ as the \emph{known classes}, and $\cY_s \setminus \cY_t$  (resp., $\cY_t \setminus \cY_s$ ) as the \emph{unknown class} in the source (resp., target) domain. The goal of open set domain adaptation is to identify known classes from the target domain, and also correctly assign labels to them.

\subsection{Self-ensemble for Domain Adaptation}
We build our model on the state-of-the-art self-ensemble model~\cite{french2018selfensembling}. It was designed for the closed set domain adaptation, where the source and target share exactly the same categories.

As shown in Fig \ref{fig:model}, in the self-ensemble model, two networks with same architecture are used: a student network, and a teacher network with its weights being automatically set as an exponential moving average of weights of the student network. The student network is trained to minimize the classification loss on labeled source samples, and also maintains consistent prediction with the teacher network for unlabeled target samples with high prediction confidence. Let us denote the student network as $f(\x)$ and the teacher network as $g(\x)$, the loss function of self-ensemble model can be generally written as:
\begin{equation}
\sum_{i=1}^{n_s} \ell_{CE}(f(\x^s_i), y^s_i) + \sum_{\x^t_i \in \cH} \left( f(\x^t_i)-g(\x^t_i)\right)^2,
\end{equation}
where $\ell_{CE}(\cdot, \cdot)$ is the cross entropy loss, $\cH$ is the set of target samples with high prediction confidence.
As explained in~\cite{french2018selfensembling}, high prediction confidence implies positive correlation between the teacher and student networks. Thus, by minimizing the square difference of two networks over those confidentially predicted target samples, the teacher network gradually guided the student network in a positive way to fit the target domain.

Recall that in open set domain adaptation problem, the source and target distribution are intrinsically different due to the unknown classes, making it unsuitable to directly minimize the distribution difference as in most traditional domain adaptation works. Thus, we propose to address the open set adaptation problem based on the teacher-student networks as in the self-ensemble model. To handle the open set issue, we design a known classes aware recognition module for identifying the known classes, and also a known classes aware adaptation module to more effectively guide the student network with the teacher network. We explain the two new modules in follows.


\subsection{Known-class Aware Recognition}
The first issue in open set domain adaptation is to identify the samples of known classes from the target domain. For convenience of presentation, we use $\cY_s^C$ and $\cY_s^U$ to denote respectively the known classes $\cY_s \cap \cY_t$ and the unknown class $\cY_s \setminus \cY_t$ in the source domain. The source domain samples of the known and unknown classes are denoted accordingly as $\cX_s^C$ and $\cX_s^U$. Similarly, the known and unknown classes and samples therein are respectively denoted as $\cY_t^C$, $\cY_t^U$, $\cX_t^C$ and $\cX_t^U$ for the target domain. Noting the fact that $\cY_s^C = \cY_t^C$, we use $\cY^C$ for simplicity. 

An intuitive way might be to learn a classifier to separate samples of known classes from those of unknown classes by using the labeled source data. However, since we have $\cY_t^U \neq \cY_s^U$, the classifier trained in the source domain can hardly guarantee to well separate  samples of known and unknown classes in the target domain. Moreover, the domain shift problem makes this issue even harder. 

To this end, we propose to identify the samples of known classes and unknown classes based on entropy measurement. Suppose that we have a classifier $f$ trained using $\cX_s^C$, then the predictions of $f$ on $\cX_s^C$ can be expected with low entropy. Meanwhile, the entropy of the predictions of $f$ on $\cX_s^U$ would be relatively high, since none of samples in $\cX_s^U$ belongs to any of the known classes. For the target domain, although samples from the unknown class $\cX^U_t$ usually belong to different classes as $\cX_s^U$, the entropy of the predictions of $f$ on $\cX^U_t$ could also be expected to be relatively high, as none of samples in $\cX_t^U$ belongs to any of the known classes. Thus, we train the model to encourage such entropy difference for distinguishing the samples from known and unknown classes.

In particular, for the student network $f(\x)$, we on one hand minimize the cross entropy loss based on $\cX_s^C$, which leading to a classifier for predicting the known classes. On the other hand, we also maximize the entropy on $\cX_s^U$, such that the entropy difference between samples of known and unknown classes would be enhanced. Therefore, the loss for the student network can be written as:
\begin{equation}
\cL_{S} = \sum_{\x_i^s\in \cX_s^C} \ell_{CE}(f(\x^s_i), y^s_i) - \sum_{\x_i^s\in \cX_s^U} \ell_{E}(f(\x^s_i)),
\end{equation}
where $\ell_{CE}$ and $\ell_{E}$ are respectively the cross entropy and entropy losses. 


\subsection{Known-class Aware Adaptation}
The second issue is to effectively guide the student network for recognizing known classes in the target domain. Although the self-ensemble model exhibits excellent performance for the closed set domain adaptation problem, samples from unknown classes in the target domain may confuse the student network if they are used by the teacher network for minimizing the square difference.  Moreover, a proper confidence threshold for selecting target samples is also crucial in the self-ensemble model, which usually needs to be carefully tuned.

To handle the above issues, we improve the self-ensemble model with a known classes aware loss for open set domain adaptation. As discussed in last section, the entropy of the predictions from the student network is helpful for distinguishing the samples of known and unknown classes. Therefore, we revise the square difference loss in the self-ensemble model with a weight term $w_i$ for each target sample $\x_i^t$. The weight $w_i$ is calculated based on the entropy of the prediction $f(\x^t_i)$ from the student network $f$ on $\x_i^t$. The higher the entropy is, the more likely $\x^t_i$ belongs to unknown classes. And thus, the value of the corresponding $w_i$ should be lower. In this case, we define the weight as:
\begin{equation}
w_i = \exp{\left(-\frac{1}{C}\sum_{k=1}^{C}f_k(\x^t_i)\log f_k(\x^t_i)\right)},
\end{equation}
where $C$ denotes the number of known classes and $f_k(\x_i)$ represents the probability of $\x_i$ being classified into the $k$-th class. Accordingly, the weighted square difference loss can be written as follows:
\begin{equation}
\cL_{U} = \sum_{i=1}^{n_t}w_i\left(f(\x_i^t) - g(\x_i^t)\right)^2.
\end{equation}

Note that, with the new known classes aware loss, the selection of confidentially predicted samples is not needed anymore, and we avoid to tune the hyper-parameter for the confidence threshold.

\subsection{Network Overview}
As shown in Fig. 1, our newly proposed KASE model inherits a similar teacher-student structure from the self-ensemble model. We first replace the original cross entropy loss with Eq. (2) for recognizing the known classes while obscuring the unknowns. Then, we use the weighted square difference loss in Eq. (4) for effectively adapt the student network to the target domain with the guidance of the teacher network. Moreover, similarly as in the self-ensemble model, data augmentation and class balance loss are also employed in our KASE model for learning a robust model. In summary, we aim to minimize the following objective in KASE:
\begin{equation}
\cL = \cL_{S} + \lambda_1\cL_{U} + \lambda_2\cL_{B},
\label{eq:overview}
\end{equation}
where $\cL_{B}$ is the class balance loss we introduce to deal with the data imbalance issue in the target domain, $\lambda_1$ and $\lambda_2$ are predefined trade-off parameters (in the experiments, we set $\lambda_1=10$ and $\lambda_2=0.1$). Similarly as in self-ensemble~\cite{french2018selfensembling}, during model training, we also compute $\cL_B$ as a cross-entropy loss between the mean probability vector and a uniform probability vector, for each mini-batch of target training samples.

\noindent\textbf{Recognizing unknown classes in the target domain}: After training KASE model, we freeze it and train a two-layer network for distinguishing known and unknown classes (\ie, binary classification) on top of KAR branch. In the test phase, we apply the learned teacher model for recognizing the target domain samples. All target samples will be assigned labels according to the prediction from KAR module except those identified as the unknown class.

\begin{table*}[t]
\centering
\small
\caption{Accuracies (\%) of different methods on the Syn2Real-O dataset.}
\vspace{-5pt}
\tabcolsep3pt
\begin{tabular}{lccccccccccccc|c}
\hline
\multicolumn{1}{l|}{Method} & plane & byc & bus & car & horse & hse & cycl & psn & plant & sktbd & train & truck & ukn & mAcc \\ \hline
\multicolumn{1}{l|}{Source Only~\cite{Peng2018Syn2RealAN}}  & 23.1 & 24.2 & 43.1 & 40.0 & 44.1 & 0.0 & 56.1 & 2.0 & 24.0 & 8.3 & 47.0 & 1.1 & \textbf{93.0} & 31.2\\
\multicolumn{1}{l|}{DAN~\cite{long2016unsupervised}} &81.3 & 76.9 & 79.5 & 68.8 & 84.0 & 32.3 & 90.5 & 44.5 & 67.8 & 41.7 & 77.8 & 5.2 & 57.8 & 62.1\\ 
\multicolumn{1}{l|}{AdaBN~\cite{adabn}} & 73.6 & 73.7 & 80.4 & 69.2 & 87.8 & 33.3 & 90.0 & 36.8 & 67.0 & 45.6 & 77.3 & 6.3 & 57.9 & 61.5 \\
\multicolumn{1}{l|}{DANN~\cite{pmlr-v37-ganin15}} &72.2 & 76.3 & 73.5 & 70.5 & 86.4  & 42.0 & 91.7 & 54.0 & 76.2 & 52.2 & \textbf{82.2} & 9.0 & 37.8 & 63.4\\

\multicolumn{1}{l|}{AODA~\cite{Saito_2018_ECCV}}  & 80.2 & 63.1 & 59.1 & 63.1 & 83.2 & 12.1 & 89.1 & 5.0 & 61.0 & 14.0 & 79.2 & 0.0 & 69.0 & 52.2\\
\hline




\hline
\multicolumn{1}{l|}{SE~\cite{french2018selfensembling}}  & \textbf{94.2} & 74.1 & 86.1 & 68.1 & \textbf{91.0} & 26.1 & \textbf{95.2} & 46.0 & \textbf{85.0} & 40.4 & 79.2 & 11.0 & 51.0 & 65.2 \\

\multicolumn{1}{l|}{Ours (w/o KAA)} & 89.8 & 82.1 & 83.6 & 64.8 & 87.8 & 46.9 & 91.0 & 65.5 & 76.7 & 54.4 & 81.8 & 15.9 & 42.9 & 67.9 \\
\multicolumn{1}{l|}{Ours}  & 89.0 & \textbf{85.6} & \textbf{88.0} & 62.7 & 89.8 & \textbf{54.1} & 90.5  & \textbf{75.8} & 81.1 & \textbf{57.5} & 79.4 & \textbf{16.8} & 41.8 & \textbf{70.2}\\ \hline
\end{tabular}
\vspace{-3mm}
\label{tab:Visda}

\end{table*}

\section{Experiments}\label{sec:experiment}
We validate our proposed KASE method for the image recognition task under the open set domain adaptation scenario. We use three benchmark datasets: Syn2real-O~\cite{Peng2018Syn2RealAN}, Office-31~\cite{Saenko_office} and Digits. 

We implement our KASE model based on the released code of the self-ensemble method\footnote{\url{https://github.com/Britefury/self-ensemble-visual-domain-adapt}}. We use three fully-connected layers with batch normalization and a ReLU activation layer after the convolutional layers as our known classification networks. 
The unknown class classification networks are implemented with two layers of fully-connected networks. 

For the setting of close set baselines, we follow the open set domain adaptation protocol in~\cite{Busto_2017_ICCV,Peng2018Syn2RealAN}, treat the unknown classes as an additation class and train a (C+1)-way classifier (C for known class, and 1 for the unknown). In the evaluation, the accuracy is obtained based on the (C+1)-way classifier on the test samples from the target domain. 

In the evaluation, we use the same evaluation metrics used in ~\cite{Busto_2017_ICCV} that all images from unknown classes in the target domain are treated as the ``unknown" class, and the mean accuracy (mAcc) is reported for comparison by averaging the accuracies of all classes including the ``unknown" class. 


\subsection{Syn2Real-O Dataset}
\label{sec:exp_visda}

\noindent{\bf Experimental setup.}
The Syn2Real-O dataset is constructed to perform object classification in real images by learning from synthetic images.  The source images were generated by rendering 3D models of 12 common classes and 33 background classes from different angles and under different lighting conditions. It contains 152,397 synthetic images. The validation set contains  55,399 images collected from Microsoft COCO dataset~\cite{mscoco}, which is used as the target domain in the experiment. 
We follow \cite{Peng2018Syn2RealAN} to adopt the ResNet-152~\cite{He_2016_resnet} model pre-trained on ImageNet as the backbone network for all methods.

\vspace{3pt}
\noindent{\bf Experimental results.}
The results of classification accuracies are summarized in Table ~\ref{tab:Visda}. The results of baseline methods ``Source Only", AODA and SE are taken from ~\cite{Peng2018Syn2RealAN}. We also additionaly report the results of conventional domain adaptation methods AdaBN, DAN, and DANN for comparison. As noted by ~\cite{Peng2018Syn2RealAN}, the class imbalance caused by the large number of images from unknown class often leads to model bias, so we apply simple reweighting strategy for all methods except AODA which does not utilize the unknown class in the source domain, \ie, assigning a weight $1/r$ to the cross-entropy loss of each class where $r$ is the ratio of this class in the source domain. We observe that such reweighting strategy generally improves those baseline methods by a large margin. The results of those methods without class balance are included in Supplementary\footnote{\url{https://bit.ly/2NoyKtt}} for reference. To further validate the effects of different components in our approach, we also report the results by removing the known-class aware adaptation (KAA) module (referred to as ``Ours (w/o KAA)"), and removing both the KAA module and the known-class aware recognition (KAR) module (referred to as ``Ours (w/o KAA and KAR)"). 
From the results, we observe that, due to the existing of lots of unknown classes in target domain, conventional domain adaptation methods as well as the source only model do not perform well on the open set domain adaptation task. In particular, they tend to incorrectly predict target samples as unknown class. The possible reason is that the samples from unknown classes are different in source and target domains, thus making the domain distribution alignment inaccurate. The open set domain adaptation method AODA does not perform well on this dataset, might because of the heavey imbalanced data in the target domain as noted in~\cite{Peng2018Syn2RealAN}. 

The special case of our method, Ours (w/o KAA and KAR), which is also the original SE model proposed in~\cite{french2018selfensembling}, performs better than the other conventional domain adaptation methods, showing the robustness of self-ensembling for handling open set domain adaptation task when compared with other distribution alignment approaches. By incorporating the known class aware recognition module (\ie, Ours (w/o KAA)), we improve the mean accuracy from $65.3\%$ to $67.9\%$.  
With our known class aware recognition (KAR) module, and known class aware adaptation module (KAA), our KASE model achieves $70.2\%$ in terms of mean accuracy, improving the naive SE model by $+4.9\%$, which validates the effectiveness of our KASE approach for open set domain adaptation.

\begin{figure}[t]
\centering
\includegraphics[width=1.0\linewidth,angle=0,trim= 0 55 0 0, clip]{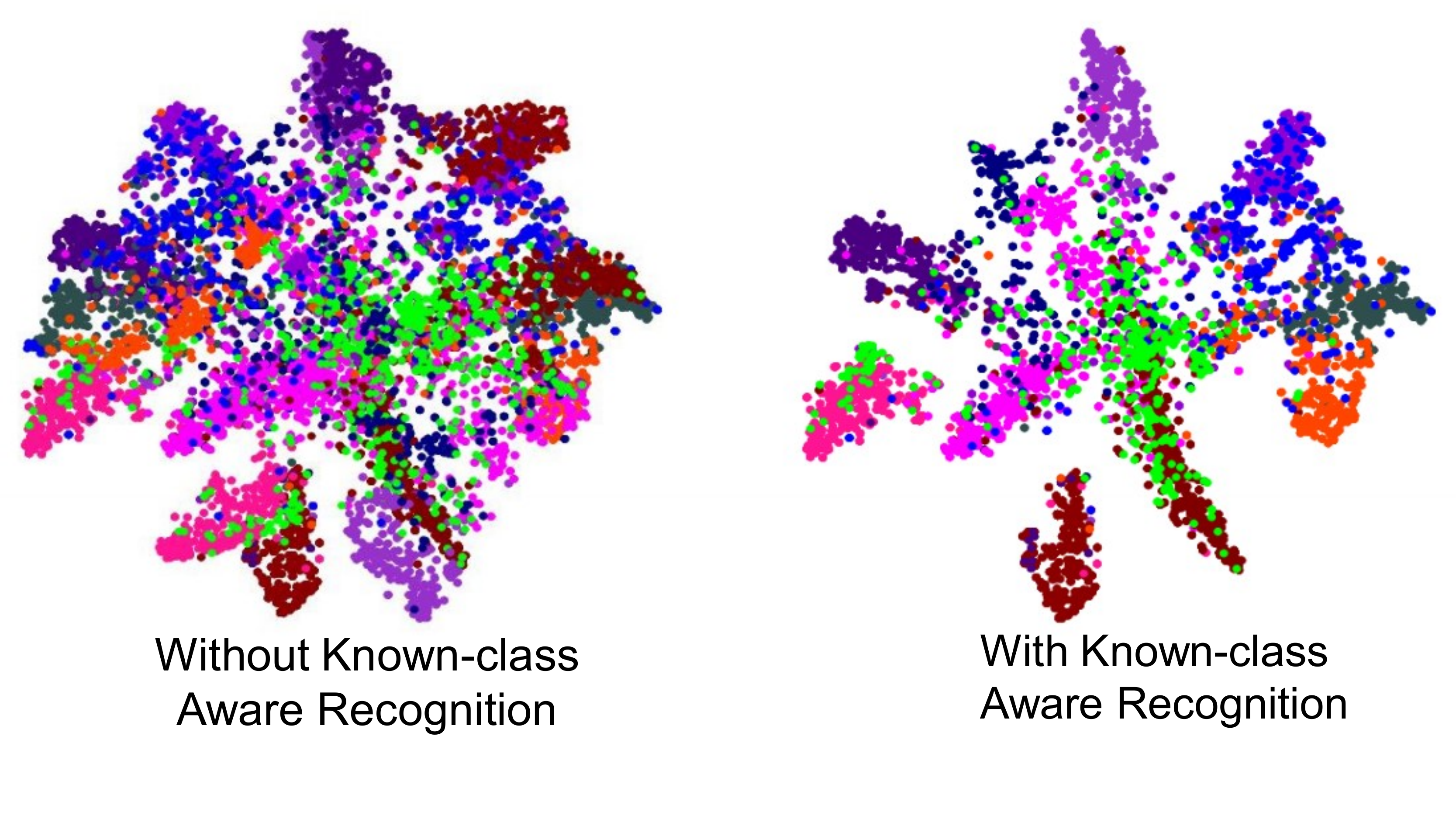} 
\vspace{-5mm}
\caption{The visualization of the feature distribution using t-SNE. We use the final convolutional layer ``res5c" to visualize. The green color denotes the unknown classes and other colors denotes known classes.}\label{fig:tsne}
\vspace{-5mm}
\end{figure}

\noindent{\bf Ablation analysis.}
To validate the effect of our proposed known-class aware recognition (KAR) module, we use t-SNE~\cite{vanDerMaaten2008} to visualize features from the last convolution layer in Fig.~\ref{fig:tsne}. We can observe that after applying KAR, the features are grouped into different clusters, and source and target domains are also well aligned.

\begin{figure}[t]
\centering
\includegraphics[width=1.0\columnwidth,angle=0, trim= 40 160 35 120, clip]{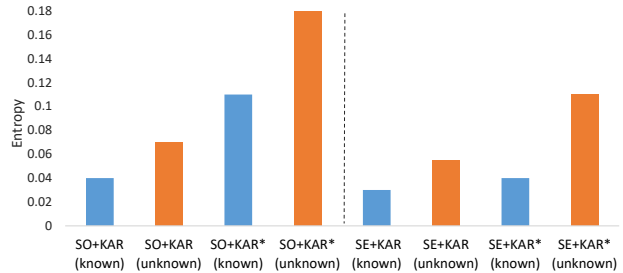} 
\vspace{-5pt}

\caption{The visualization of the entropy between known and unknown class in Syn2Real-O dataset. SO and SE mean source only and the self ensemble method, respectively. KAR means our known class aware recognition module. The method that marked with * are applied with a maximize entropy operation for the unknown classes when training.}\label{fig:entropy}
\vspace{-5pt}
\end{figure}

Furthermore, we also validate the effectiveness of using entropy information to guide the known class aware adaptation. In Fig.~\ref{fig:entropy}, we compare the entropy of known and unknown classes in the target domain using Source only and SE models combining with our KAR, respectively. We observe that, being combined with the SE model, our KAR can effectively maximize the entropy of unknown class  while successfully keeping the entropy of known classes to be relatively low, thus helping to distinguish the unknown class from known classes.

\noindent{\bf Results in VisDA Challenge 2018.}
Using the proposed KASE model, we secured the second place in the VisDA open set domain adaptation challenge 2018. Our single model based on ResNet152 yields a mean accuracy of 68.2\% on the test set. By ensembling three models with  different backbones (\ie, ResNet101, ResNet152 and SE-ResNeXt-101~\cite{hu2018senet}), we finally achieved 69.0\% in terms of mean accuracy on the test set.

\begin{table*}[t]
\centering
\small
\caption{Mean accuracies (\%) of different methods on the Office-31 dataset. AVG represents the averaged value of mean accuracies over different settings.}
\vspace{-10pt}
\begin{tabular}{l|cccccc|c}
\hline
Method & A$\rightarrow$D & A$\rightarrow$W & D$\rightarrow$A& D$\rightarrow$W & W$\rightarrow$A & W$\rightarrow$D & AVG\\ \hline
Source only & 70.3 & 60.1  & 53.4 & 86.6 & 44.8 & 90.6 & 67.6 \\    
DAN~\cite{dan} & 77.6 & 72.5 & 57.0 & 88.4 & 60.8 & 98.3 & 75.7\\
DANN~\cite{pmlr-v37-ganin15} & 78.3 & 75.9 & 57.6 & 89.8 & 64.0 & \textbf{98.7} & 77.4 \\
AIT~\cite{Busto_2017_ICCV} & 79.8 & 77.6 & 71.3 & 93.5 & 76.7 & 98.3 & 82.9 \\ 
AODA~\cite{Saito_2018_ECCV} & 76.6 & 74.9 & 62.5 & 94.4 & 81.4 & 96.8 & 81.1\\
D-FRODA~\cite{baktashmotlagh2018learning} &\textbf{87.4} & 78.1 & 73.6 & 94.4 & 77.1 & 98.5 & 84.9 \\ 
\hline
SE~\cite{french2018selfensembling} & 74.3 & 73.3 & 58.0 & 93.4 & 63.6 & 91.4 & 75.6\\ 
Ours  & 87.0 & \textbf{80.3} & \textbf{78.0} & \textbf{95.4} & \textbf{81.8} & 98.6 & \textbf{86.9} \\ \hline
\end{tabular}
\vspace{-15pt}    
\label{tab:office}
\end{table*}

\subsection{Office-31 Dataset}
We further evaluate our KASE model on the benchmark Office-31 dataset~\cite{Saenko_office}.
It consists of three domains: Amazon (A), DSLR (D) and Webcam (W), each of which contains images from 31 common classes. The Amazon dataset contains centred object on clean background, while the other two are taken in an office environment but with different cameras. 
We follow the experiment protocol in~\cite{Busto_2017_ICCV} for conducting open set domain adaptation. The 10 common classes with the Office-Caltech dataset~\cite{gong_gfk} are treated as the known classes. Then, the classes 11-20 are used as unknown classes in the source domain, and the classes 21-31 as unknown classes in the target domain. By using one domain as the source and another as the target, we obtain 6 cases, \ie,  4 with a considerable domain shift (A $\rightarrow$ D, A $\rightarrow$ W, D $\rightarrow$ A, W $\rightarrow$ A) and 2 with minor domain shift (D $\rightarrow$ W, W $\rightarrow$ D). Following \cite{Busto_2017_ICCV}, AlexNet~\cite{AlexNet} is used as our backbone. The other settings are the same as in Section~\ref{sec:exp_visda}.


The experimental results are shown in Table~\ref{tab:office}. We compare our KASE model with existing state-of-the-arts in both open set and closed set approaches~\cite{dan,pmlr-v37-ganin15,french2018selfensembling,Busto_2017_ICCV,Saito_2018_ECCV,baktashmotlagh2018learning}. The results of DAN and DANN are taken from \cite{Busto_2017_ICCV}, and results of ATI, AODA and D-FRODA are from their original papers. Similar as in the Syn2Real-O experiment, we observe that conventional distribution alignment method~\cite{dan,pmlr-v37-ganin15} do not perform well, especially for cases with large domain shift and scarce source domain (\eg, D$\rightarrow$A, W$\rightarrow$A), while the open set domain adaptation methods AIT, AODA and  D-FRODA perform better. Our proposed KASE model achieves the best performance, which again proves the effectiveness of our approach for open set domain adaptation. 

\begin{table}[t]
    \begin{center}
    \setlength{\tabcolsep}{3pt}
    \small
    \caption{Mean accuracies (\%) of different methods on the Digits Dataset (S: SVHN, M: MNIST, U: USPS). ``AVG" represents the averaged value of mean accuracies over different settings.}
    \begin{tabular}{l|ccc|c}
    \hline
    Method& S-M & U-M & M-U & AVG \\ \hline   
    Source only & 60.0 & 80.5 & 80.8 & 73.8\\
    DAN~\cite{long2016unsupervised} & 65.6 & 86.3 &  88.6 & 80.2 \\
    DANN~\cite{pmlr-v37-ganin15} & 65.3 & 88.4 & 87.3 & 80.3  \\
    AODA~\cite{Saito_2018_ECCV} & 63.0 & 93.2 & 92.4 & 82.8 \\
    \hline
    SE~\cite{french2018selfensembling} & 63.4 & 93.0 & 90.6 & 82.3 \\    
    Ours & \textbf{66.8} & \textbf{94.4} & \textbf{93.1} & \textbf{84.8}\\
    \hline
    \end{tabular}
    \label{tab:digit}
    \end{center}
\vspace{-4mm}    
\end{table}

\subsection{Digits Dataset}
Following the previous work~\cite{Saito_2018_ECCV}, we also evaluate our proposed KASE model on the Digits dataset under the open set domain adaptation scenario. 
Three cases are considered: SVHN to MNIST, USPS to MNIST and MNIST to USPS. We set 0 to 3 as known categories 4-6 as source unknown categories and 7-9 as target unknown categories. The same backbone as in the ~\cite{Saito_2018_ECCV}, and the other settings are the same as in Section~\ref{sec:exp_visda}. The results are reported in Table~\ref{tab:digit}, where our KASE outperforms existing state-of-the-art in both open set and closed set approach~\cite{dan,pmlr-v37-ganin15,french2018selfensembling,Saito_2018_ECCV}, which again demonstrates the superiority of our proposed approach for open set domain adaptation.



\vspace{-3pt}
\section{Conclusion}\label{sec:conclusion}
\vspace{-2pt}
In this paper, we have proposed a new method called known class aware self-ensemble (KASE) for open set domain adaptation. To handle the challenges caused by different labels space, we designed two modules to effectively identify known and unknown classes and perform domain adaptation based on the likeliness of target samples belonging to known classes. We implemented our approach based on the recent self-ensemble model, in which the two new modules are trained jointly in an end-to-end fashion. Extensive experiments on multiple benchmark datasets have demonstrated the superiority of our KASE model compared to existing state-of-the-art domain adaptation methods. 


\bibliographystyle{named}
\bibliography{ijcai19}

\end{document}


\maketitle
In this Supplementary Material, we provide some details omitted in the main text.
\begin{itemize}
    \item Section ~\ref{sec:visda}: The full experiments results in the Syn2Real-O dataset
\end{itemize}

\section{The full experiments results in the Syn2Real-O dataset}
~\label{sec:visda}

\begin{table*}[hb!]
\centering
\small
\caption{The classification results on the Syn2Real-O. Results of ``Source Only", AODA, and SE are taken from ~\cite{Peng2018Syn2RealAN}.}
\vspace{-10pt}
\tabcolsep3pt
\begin{tabular}{lcccccccccccccc}
\hline
\multicolumn{1}{l|}{Method} & plane & byc & bus & car & horse & hse & cycl & psn & plant & sktbd & train & truck & ukn & mAcc \\ \hline
\multicolumn{1}{l|}{Source Only~\cite{Peng2018Syn2RealAN}}  & 23.1 & 24.2 & 43.1 & 40.0 & 44.1 & 0.0 & 56.1 & 2.0 & 24.0 & 8.3 & 47.0 & 1.1 & \textbf{93.0} & 31.2\\
\multicolumn{1}{l|}{DAN~\cite{long2016unsupervised}} &81.3 & 76.9 & 79.5 & 68.8 & 84.0 & 32.3 & 90.5 & 44.5 & 67.8 & 41.7 & 77.8 & 5.2 & 57.8 & 62.1\\ 
\multicolumn{1}{l|}{DAN (w/o class balance)} & 70.6 & 65.9 & 73.5 & 63.8 & 80.8 & 17.9 & 83.1 & 16.3 & 26.0 & 31.1 & 75.9 & 5.5 & 88.6 & 53.8\\
\multicolumn{1}{l|}{AdaBN~\cite{adabn}} & 74.5 & 63.7 & 77.0 & 63.9 & 78.3 & 24.2 & 89.1 & 38.0 & 33.9 & 39.0 & 75.4 & 5.6 & 69.3 & 56.3 \\
\multicolumn{1}{l|}{AdaBN(w/o class balance)} &72.3 & 70.3 & 77.2 & 65.6 & 83.3 & 8.6 & 84.3 & 21.3 & 31.3 & 28.5 & 67.8 & 7.4 & 85.6 &  54.1\\
\multicolumn{1}{l|}{DANN~\cite{pmlr-v37-ganin15}} &72.2 & 76.3 & 73.5 & 70.5 & 86.4  & 42.0 & 91.7 & 54.0 & 76.2 & 52.2 & 82.2 & 9.0 & 37.8 & 63.4\\
\multicolumn{1}{l|}{DANN (w/o class balance)} & 70.3 & 73.4 & 80.8 & 67.0 & 85.1 & 21.3 & 84.8 & 32.3 & 52.6 & 34.2 & 71.1 & 8.5 & 77.4 & 58.4 \\
\multicolumn{1}{l|}{SE~\cite{french2018selfensembling}}  & \textbf{94.2} & 74.1 & 86.1 & 68.1 & \textbf{91.0} & 26.1 & \textbf{95.2} & 46.0 & \textbf{85.0} & 40.4 & 79.2 & 11.0 & 51.0 & 65.2 \\
\multicolumn{1}{l|}{AODA~\cite{Saito_2018_ECCV}}  & 80.2 & 63.1 & 59.1 & 63.1 & 83.2 & 12.1 & 89.1 & 5.0 & 61.0 & 14.0 & 79.2 & 0.0 & 69.0 & 52.2\\
\hline



\hline
\multicolumn{1}{l|}{Ours (w/o KAR and KAA)}  & 90.2 & 78.1 & 84.9 & \textbf{75.4} & 90.3 & 25.1 & 94.0 & 51.3 & 76.2 & 38.1 & 73.3 & 9.8 & 62.5 & 65.3\\

\multicolumn{1}{l|}{Ours (w/o KAA)} & 89.8 & 82.1 & 83.6 & 64.8 & 87.8 & 46.9 & 91.0 & 65.5 & 76.7 & 54.4 & ~\textbf{81.8} & 15.9 & 42.9 & 67.9 \\
\multicolumn{1}{l|}{Ours}  & 89.0 & \textbf{85.6} & \textbf{88.0} & 62.7 & 89.8 & \textbf{54.1} & 90.5  & \textbf{75.8} & 81.1 & \textbf{57.5} & 79.4 & \textbf{16.8} & 41.8 & \textbf{70.2}\\ \hline
\end{tabular}
\vspace{-5pt}
\label{tab:Visda}

\end{table*}



As discussed in the main paper section 4, we apply a simple reweighting strategy for serveral baseline methods (\ie, DAN, AdaBN, DANN) to solve the class imbalance problem in the source domain. Here, we present the experiment results without using this operation in the table ~\ref{tab:Visda}. We could find that the mAcc gain a lot after applying this strategy for all baseline methods.

\bibliographystyle{named}
\bibliography{ijcai19}